\pdfoutput=1
\documentclass[11pt]{article}
\usepackage[preprint]{acl}
\usepackage{times}
\usepackage{latexsym}
\usepackage[T1]{fontenc}
\usepackage[utf8]{inputenc}
\usepackage{microtype}
\usepackage{inconsolata}
\usepackage{graphicx}
\usepackage{booktabs}
\usepackage{amsmath}
\usepackage{xurl}

\title{Do Earnings Calls Help Predict Financial Restatements?\\
Evaluating Text Beyond Public Firm Information}

\author{Ahmet Efe, Shivank Sapra, Bilal Ahmed \\
  University of Minnesota, Minneapolis, MN, USA \\
  \texttt{\{efe00002, sapra013, ahme0692\}@umn.edu}}

\hypersetup{pdftitle={Do Earnings Calls Help Predict Financial Restatements? Evaluating Text Beyond Public Firm Information},pdfauthor={Ahmet Efe, Shivank Sapra, Bilal Ahmed}}

\begin{document}
\maketitle

\begin{abstract}
Financial restatements correct errors in previously published financial reports. Could managers' language in earnings calls warn of them in advance? Comparing 109 calls that precede restatements with 218 matched control calls, we find that the text predicts restatements but adds no detectable ranking gain beyond public firm information. A text classifier beats chance, largely because of restatements tied to special-purpose acquisition companies, but a model using only public information scores slightly higher. With refitting, the 95\% interval for the gain from adding text ($-$0.05 to 0.11 AUROC) still leaves room for a meaningful benefit. A large language model scores higher, but we show that it names the exact ticker in 66\% of masked transcripts, and it was released after the restatements occurred. Our results suggest that finding new warning signs in call text requires a public-information baseline and a check of what a language model already knows.
\end{abstract}

\section{Introduction}
Investors rely on financial reports to understand a company's performance and position. When errors are discovered in previously published reports, a company may need to correct them through a financial restatement. These corrections can be costly for investors \citep{palmrose2004determinants}, motivating research on whether warning signs can be detected before restatements are announced. Previous studies have examined information in financial statements \citep{dechow2011predicting,bao2020detecting} and corporate disclosures \citep{humpherys2011identification,purda2015accounting,brown2020topic}.

Earnings calls offer another possible source of warning signs. Executives discuss the company's results in prepared remarks and then answer questions from financial analysts. Prior research links executives' language in these calls to later restatements \citep{larcker2012deceptive}. Most restatements, however, correct errors rather than fraud \citep{hennes2008errors}, and managers may not yet know that a correction will be needed.

A central challenge is understanding what a successful text model has learned. Firms that later restate may already differ from other firms in publicly observable ways, such as their size or profitability. Their calls may reflect these characteristics, so a model can learn to recognize risky firms rather than an approaching restatement. This is a form of shortcut learning \citep{geirhos2020shortcut}, because firm characteristics can shape both the language of a call and the likelihood of a later restatement \citep{keith2020text}. Measuring what text adds therefore requires a baseline built from public information.

We construct a new dataset of 109 calls that precede restatement disclosures, each linked to its SEC filing and matched to two control calls from firms without an observed restatement. We find that a baseline built from public information scores slightly higher than the text, and that adding text features yields no detectable ranking gain. Much of the text signal comes from restatements tied to special-purpose acquisition companies. The signal shows no consistent trend as the disclosure approaches, and the pre-filing call does not stand out from the firm's own earlier calls. Finally, a large language model recognizes most masked firms and does nearly as well from the name alone.

Our contributions are:
\setlength\leftmargini{1.2em}
\begin{itemize}\setlength\itemsep{0pt}
\item A new matched case--control dataset of earnings calls before restatements.
\item We show that public information alone scores slightly higher than a text model that beats chance, and that adding text yields no detectable gain in any of the four settings.
\item We show that a large language model names the exact ticker in 66\% of masked calls, so masking does not anonymize firms.
\end{itemize}

\section{Related Work}
Earnings-call research has used language to predict market outcomes such as volatility and earnings surprises \citep{qin2019say,koval2023forecasting}, and to study analyst behavior \citep{keith2019modeling}. Other work examines how executives communicate under scrutiny. \citet{larcker2012deceptive} relate executives' language to later restatements, \citet{hobson2012speech} study vocal cues, and research on non-answers \citep{gow2021nonanswers} and evasive responses \citep{nuaimi2025evasive} asks whether analyst Q\&A reveals problems. Research on financial misreporting also uses financial variables \citep{dechow2011predicting,bao2020detecting}, disclosure language \citep{humpherys2011identification,purda2015accounting}, textual topics \citep{brown2020topic}, and changes in a firm's disclosures over time \citep{cohen2020lazy}. We add a matched comparison showing that adding call text yields no detectable ranking gain over public firm information.

Language models applied to historical events may draw on information learned after the event. This concern has been raised for financial prediction \citep{lopezlira2023chatgpt,glasserman2023lookahead} and for evaluation data more broadly \citep{sainz2023contamination}. \citet{glasserman2023lookahead} anonymize company names to reduce this bias. We show that masking does not prevent firm recognition, and that the firm's name alone nearly matches the model's full-transcript score.

\section{Data}
\paragraph{Restatement cases.}
We build the dataset by linking SEC restatement filings to earnings-call transcripts. Companies report restatements to the U.S. Securities and Exchange Commission (SEC) in Form 8-K Item~4.02, which concerns non-reliance on previously issued financial statements. Our initial collection contains 8{,}903 such filings from 2005 to 2024. Earnings-call transcripts come from two public corpora: transcripts published by The Motley Fool \citep{tpotterer2023motleyfool} and a collection covering S\&P~500 firms \citep{kurry2025sp500}. We link calls to firms through the SEC's permanent company identifier rather than ticker symbols, which different firms can use at different times. For each filing, we select the firm's closest earnings call 8 to 365 days before it. We skip the final week because companies must generally file within four business days of deciding to restate. We keep only filings whose text explicitly states non-reliance, and we skip calls within a year after an earlier Item~4.02 filing by the same firm. Most filings drop out because the firm has no transcript in the prior year (156 remain, including one corpus later dropped; Appendix~\ref{app:extra}). This gives 109 case calls from 106 firms.

\paragraph{Matched controls.}
Each case is matched to two control calls from firms with no Item~4.02 filing observed in EDGAR through 2025. Controls come from the same transcript corpus and calendar quarter as the case and, for 192 of the 218 controls, the same two-digit industry. Each control firm is used only once. The final dataset contains 109 matched sets and 327 calls.

\section{Evaluation}
\paragraph{Comparing a case with its own controls.}
We ask whether a model gives a restatement case a higher risk score than its matched controls. Our metric, \emph{within-set AUROC}, measures this ordering within each matched set and averages across sets; 0.5 is chance. We use 5-fold cross-validation repeated 10 times. The folds keep each matched set together and group all sets involving the same firm, so no firm appears in both training and test data. Confidence intervals come from resampling matched sets. For the main comparison we also refit the models on each resample, which gives a wider and more realistic interval.

\paragraph{Public-information baseline.}
The baseline does not read the transcript. It uses two public warning signs filed in the year before the call, a notice that a periodic report will be late and a change of auditor, together with firm characteristics such as size, revenue, losses, growth, firm age, industry, and a flag for firms that may have listed through a shell-company merger. Financial values come from the SEC's structured filings, using the values originally reported before the call. The baseline is a logistic regression.

\paragraph{Measuring what text adds.}
Our main text model is a logistic regression on TF-IDF features of the full transcript. To measure what text adds, we fit one logistic regression on the text and baseline features together and compare it with the baseline on the same folds. We call the difference in within-set AUROC $\Delta$. The regularization strength is fixed rather than tuned. We also evaluate FinBERT tone features \citep{araci2019finbert} and text embeddings \citep{qwen3embedding} without fine-tuning, and two language models with 9 and 27 billion parameters \citep{qwen35models} without task-specific training. The language models read the full transcript with company names, tickers, and person names masked and are asked whether a restatement will follow. Details are in Appendix~\ref{app:extra}.

\paragraph{Decision rule.}
Small samples produce wide intervals, so we recorded decision rules before each stage of analysis. We consider a meaningful effect in the predicted direction ruled out when its 95\% upper bound is below 0.60 AUROC, or below 0.05 for an improvement over a baseline. Otherwise, we report the result as inconclusive. The comparison with public information and the later analyses were planned after the first text result had been seen, and two checks were added without a plan (Appendix~\ref{app:extra}). We therefore treat them as exploratory and do not correct for multiple comparisons.

\section{Results}

\begin{figure}[t]
\centering
\includegraphics[width=\linewidth]{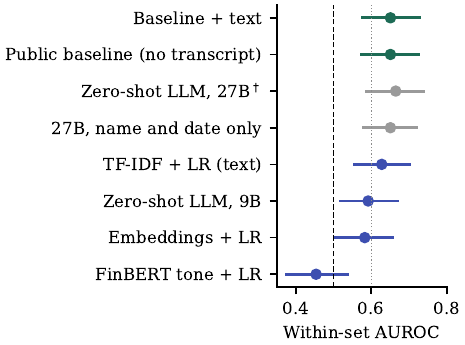}
\caption{Within-set AUROC on 109 matched sets with 95\% confidence intervals. The dashed line marks chance, and the dotted line marks the 0.60 threshold. The name-only bar gives the 27B model only the firm's name and the call date. $^\dagger$The 27B model names the exact ticker in most masked transcripts (see the language-model results).}
\label{fig:main}
\end{figure}

\paragraph{Call text predicts restatements better than chance.}
The TF-IDF model reaches a within-set AUROC of 0.628 [0.555, 0.702] (Figure~\ref{fig:main}). A permutation test that shuffles labels within matched sets gives $p=0.029$. The embedding model (0.583) and FinBERT tone (0.454) score lower. The signal appears stronger in 2015--2024 than in 2005--2014 (0.685 versus 0.464), but this difference is confounded with corpus and uncertain within one corpus. It also weakens when cases related to special-purpose acquisition companies (SPACs) are removed: a model retrained on the remaining 80 sets reaches 0.562 [0.48, 0.64], and the original model scores 0.594 on those same sets. Without these cases, the text model is no longer clearly better than chance. Many of these cases are 2021 warrant restatements around an SEC staff statement on warrant accounting. Restatements of warrant or equity classification are the easiest to rank (0.82 on 17 sets). Our results show that the text signal is above chance but comes largely from SPAC-related restatements.

\paragraph{Public information scores slightly higher than the text model.}
The public-information baseline reaches 0.651 [0.573, 0.725], slightly above the text. The text model shows no detectable advantage over the baseline ($-0.02$ [$-0.12$, 0.07], fixed models). The two filing-based warning signs alone reach only 0.548, so most of the baseline's strength comes from firm characteristics. Adding the accounting-based misstatement score of \citet{dechow2011predicting} leaves the baseline essentially unchanged ($-0.005$). Together with the next result, this shows that text adds no detectable ranking gain over public firm information.

\begin{table}[t]
\centering\small
\setlength{\tabcolsep}{2pt}
\begin{tabular}{lccc}
\toprule
Setting (sets) & Baseline & Text & $\Delta$ [95\% CI] \\
\midrule
Main (109) & 0.651 & 0.628 & $0.000$ [$-$.06, .06] \\
$\pm$3-yr controls (109) & 0.683 & 0.665 & $+0.005$ [$-$.05, .06] \\
No SPAC-related (80) & 0.625 & 0.562 & $-0.025$ [$-$.08, .03] \\
Out-of-time (69) & 0.594 & 0.572 & $+0.014$ [$-$.03, .06] \\
\bottomrule
\end{tabular}
\caption{Within-set AUROC of the baseline and the text model, and the change $\Delta$ when text is added to the baseline. Intervals hold the models fixed. $\pm$3-yr: controls with no Item~4.02 observed within three years of the call. No SPAC-related: models retrained without SPAC-related cases. Out-of-time: trained on the 40 sets filed by 2019 and tested on the later sets. Definitions are in Appendix~\ref{app:extra}.}
\label{tab:robust}
\end{table}

\paragraph{Adding text yields no detectable gain.}
We find no detectable gain, meaning an interval that includes zero, from adding the text features in any of the four settings or with either combination method. Adding them to the baseline leaves its within-set AUROC unchanged ($\Delta=0.000$), and adding the out-of-fold text score as a single feature, as we do for the language model below, gives $+0.009$ [$-0.05$, 0.06]. The estimate stays close to zero with an alternative control definition, without SPAC-related cases, and when models trained on earlier years are tested on later ones (Table~\ref{tab:robust}). By our decision rule these results are inconclusive. The upper bounds exceed 0.05 except without SPAC-related cases (0.03, models held fixed), and refitting widens the main interval to $[-0.05, 0.11]$.

A conditional logistic regression shows that the text score keeps a modest association of its own. Within matched sets, a higher text score is still associated with being a case after accounting for the baseline score (odds ratio 1.40 [1.05, 1.87] per standard deviation). The two results are compatible: the odds ratio reflects how far apart the scores are, while AUROC only counts which score is higher. When firm characteristics replace the baseline score, $p$ ranges from 0.036 to 0.075 across samples and specifications (Appendix~\ref{app:extra}).

\paragraph{Masking does not anonymize firms for a large language model.}
The 9B language model reaches 0.592 and the 27B model 0.665. Adding the 27B score to the baseline as one more feature changes it by $+0.069$ [0.01, 0.13], the only text-based gain whose fixed-model interval excludes zero. However, we find that masking does not hide the firm: asked which company a masked transcript belongs to, the 27B model names the exact ticker for 66\% of the transcripts. It recognizes controls more often than cases (71\% versus 57\%), so how well known a firm is may itself carry signal. Across the 217 exact-ticker matches, the ticker is absent from the masked input, so these answers cannot be explained by copying it. Given only the firm's name and the date, the 27B model reaches 0.651, close to its score with the full transcript. When this name-only score is also added to the baseline, the transcript still adds $+0.060$ [$-0.01$, 0.13] (fixed models). The name-only score therefore explains little of the point estimate (0.069 versus 0.060), although the remaining interval includes zero, and it cannot remove what the model recognizes from the content of the call itself. Because the model was released in 2026, after the restatements we study, it may also know their outcomes. Its higher score therefore cannot be credited to the transcript alone.

\paragraph{No consistent sign of language that anticipates the restatement.}
If managers' language anticipated a restatement, the signal might grow stronger as the announcement approaches, differ from the firm's earlier calls, or depend on explicit mentions of trouble. We find no consistent trend over time. Calls two to three years before the filing cannot be distinguished from the pre-filing call (pre-filing minus earlier, 0.01 [$-0.14$, 0.17] on 36 sets). In 98 sets, however, the case call itself presents results that were later restated, so the time to disclosure is not the time since the accounting problem began. Removing passages that mention investigations, audit-committee reviews, or control weaknesses changes the text model's AUROC by $-0.014$ [$-0.04$, 0.01], and masking names changes it by $-0.018$ [$-0.06$, 0.03]. A retrospectively fitted score of how unusual a call is relative to the firm's own earlier calls does not separate cases from controls at the pre-filing call (0.439 [0.33, 0.57] on the 49 sets with enough history), which rules out an effect of 0.60 or more. None of the planned analyses points to language that anticipates the restatement. An exploratory comparison at earlier calls and an evasiveness analysis are reported in Appendix~\ref{app:extra}.

\section{Discussion}
Our results show that a text model can predict restatements without a detectable ranking gain over public data. Call text distinguishes calls that precede restatements from matched controls, largely because of SPAC-related cases, but public filings and firm characteristics score slightly higher, and adding the text features yields no detectable gain. We interpret this as the text partly reflecting persistent differences between firms. The same firms' calls two to three years earlier score about as high as the pre-filing call (0.57 versus 0.58). Consistent with this, the text and baseline scores are correlated within matched sets (rank correlation 0.38 [0.25, 0.49]), although the baseline's firm characteristics predict only a small part of the text score ($R^2=0.05$; Appendix~\ref{app:extra}). The terms most associated with cases include investigation, audit, and young-industry words such as cannabis. The conditional analysis still finds a modest residual association, so text may carry information that rankings do not reveal. Larger corpora and language models trained on pre-event data would sharpen these tests.

These findings have three implications for financial NLP. First, text models should be compared with a baseline built from public information available at prediction time, and the gain over that baseline should be reported with its uncertainty. Second, cases should be compared with matched controls chosen by clearly stated rules. Third, when language models are evaluated on historical events, masking names is not enough: reporting how often the model recognizes the firm, and how well it predicts from the name alone, shows what its score may already contain.

\clearpage
\section*{Limitations}
Restatements are relatively rare, and only some restating firms appear in public earnings-call corpora. Together with our eligibility rules, this limits the sample to 109 matched sets, and one corpus focuses on S\&P~500 firms. Our conclusions about the value of text refer to the models and combination methods evaluated here, and other combinations may extract more information. The annotations used in secondary analyses come from language models and were checked for consistency with independent model raters. Finally, restatements frequently stem from accounting errors rather than intentional misconduct, so our predictions should not be read as evidence of wrongdoing.

\section*{Ethics Statement}
Predictions of restatement risk should not be treated as evidence of fraud or deliberate wrongdoing. Misleading interpretations could cause reputational harm to companies and individuals, and the evaluation supports research conclusions rather than decisions about individual firms. We will not redistribute transcripts, because redistribution rights vary across sources.

\section*{Acknowledgments}
The authors acknowledge the Minnesota Supercomputing Institute (MSI) at the University of Minnesota for providing resources that contributed to the research results reported in this paper.

\section*{Use of AI Assistants}
AI assistants supported coding, data auditing, and manuscript editing. Language models were also used for annotation and scoring, as described in the methods and results. The authors reviewed the analyses and claims and take responsibility for the final manuscript.

\bibliography{custom}

\appendix

\section{Additional Details and Results}\label{app:extra}

\paragraph{Sample construction.}
We remove duplicate transcripts of the same call. A third public corpus was not used because it contains only partial transcripts. Company names in transcripts are checked against SEC records. We also exclude calls made within one year after an earlier Item~4.02 filing by the same firm. Of the 8{,}903 filings, 156 have a transcript of the firm in the year before the filing in any of three corpora, 143 have a call whose company name matches SEC records, 139 have such a call more than seven days before the filing, and 134 remain after the one-year rule. These give 123 unique case calls, 111 after removing the partial-transcript corpus, and 109 after removing filings whose text does not explicitly state non-reliance. Of the 109 matched sets, 28 are from 2005--2014 and 81 from 2015--2024. Controls are aligned in time with their case: each control call falls in the same calendar quarter, and time-based analyses use the case's filing date for the whole set.

\paragraph{Dataset summary.}
Table~\ref{tab:data} summarizes the dataset. Case filings cluster in 2020--2023 (65 of 109 sets), largely because of the 2021 wave of warrant restatements. The most common case industries are business services (17 sets) and electrical and electronic equipment (14 sets). The pre-filing call precedes the filing by a median of 91 days.

\begin{table}[t]
\centering
\small
\begin{tabular}{lrr}
\toprule
 & Cases & Controls \\
\midrule
Calls & 109 & 218 \\
Distinct firms & 106 & 218 \\
Motley Fool corpus & 61 & 122 \\
S\&P~500 corpus & 48 & 96 \\
Calls in 2005--2014 & 28 & 56 \\
Median words & 8{,}308 & 8{,}682 \\
Interquartile range & 6{,}800--9{,}469 & 7{,}282--9{,}741 \\
Calls with separate Q\&A & 104 & 205 \\
\bottomrule
\end{tabular}
\caption{Dataset summary. Each control matches its case's corpus and quarter.}
\label{tab:data}
\end{table}

\paragraph{Baseline details.}
Warning signs are counted when filed between one day and one year before the call. Auditor changes filed together with a completed merger or a change in shell-company status are excluded. The baseline also includes dividends, share repurchases, public float, exchange, and era. The shell-company flag uses the firm's name, industry code, or a filing history shorter than three years. Structured financial data are available mainly from 2010 onward. Missing values are imputed from the training folds and marked with indicator variables.

\paragraph{Evaluation details.}
In each of the 10 repetitions, out-of-fold scores are converted to ranks divided by the test-fold size and then averaged across repetitions. Where a logit of these scores is needed, they are clipped to $[10^{-4}, 1-10^{-4}]$. Within-set AUROC counts ties as one half. Fixed-model intervals use 2{,}000 bootstrap resamples of matched sets. The refitted interval for the main comparison uses 1{,}000 resamples and refits the vectorizer, imputation, scaling, and classifier each time; a set drawn twice stays within one fold, so the copies never appear in both training and test data. The refitted resamples have a mean $\Delta$ of 0.024, which shifts the refitted interval upward relative to the fixed-model interval. Resampling case firms with all their sets gives nearly the same fixed-model interval for the main comparison ($[-0.06, 0.05]$ instead of $[-0.06, 0.06]$). The permutation test shuffles labels within sets, refits the whole pipeline, and uses 1{,}000 permutations on one repetition of the folds.

\paragraph{Model details.}
TF-IDF uses English stop words, alphabetic tokens of at least two letters, a minimum document frequency of 2, a maximum document frequency of 0.95, and sublinear term frequency. Logistic regression uses $C=1$, balanced class weights, and the liblinear solver. In the combined model, the standardized baseline features are multiplied by $1/\sqrt{k}$, where $k$ is the number of baseline features, so that the two feature groups have similar overall magnitude. Regularization is fixed rather than tuned. The conditional logistic regression uses the logit of each model's averaged rank, standardized over all calls, so its odds ratio is per standard deviation of that score. Models are pinned to Hugging Face revisions: \texttt{Qwen/Qwen3.5-9B} \texttt{c2022362}, \texttt{Qwen/Qwen3.5-27B} \texttt{fc05daec}, \texttt{Qwen/Qwen3-Embedding-8B} \texttt{1d8ad4ca}, and \texttt{ProsusAI/finbert} \texttt{4556d130}. The Qwen3.5 models read each transcript in full (at most 30{,}000 tokens; none was truncated). FinBERT labels each sentence (at most 512 tokens), and its features are the shares of positive, negative, and neutral sentences. Embeddings average non-overlapping 8{,}192-token chunks.

\paragraph{Prompts and firm recognition.}
For prediction, the language models received the masked transcript and the question ``Within the next twelve months, will this company disclose that its previously issued financial statements should no longer be relied upon (a restatement)? Answer Yes or No.'' The score is $P(\text{Yes})/(P(\text{Yes})+P(\text{No}))$ from the next-token distribution. For recognition, each model (9B and 27B) was asked to name the company and ticker of the masked transcript. Over the 327 masked transcripts, the ticker named by the 27B model equals the firm's ticker in 66\% (57\% of cases and 71\% of controls). Exact matching may miss valid identifications when the model gives an earlier ticker of a firm that later changed it. In none of the 217 exact matches does the ticker appear in the masked transcript, so these identifications come from the content of the call. A looser rule, counting an answer as correct when it contains the ticker or the first word of the firm's SEC name, gives 75\%, and 54\% for the 9B model. Under this rule, stronger masking that also removes every capitalized word, number, and year still matches the firm in 63\% of transcripts for the 27B model, and its gain over the baseline under that masking is $+0.041$ [$-0.02$, 0.11]. Under the looser rule, the 27B model's within-set AUROC is 0.69 [0.58, 0.79] on the 70 sets where it recognizes the case and at least one control, and 0.73 [0.50, 0.92] on the 13 sets where it does not recognize the case and also misses at least one control; the other 26 sets fall in neither group. For the 27B model, using the first distinctive word of the name instead, after removing generic words such as ``First'' or ``American'', gives 77\%. The name-only score asks the same restatement question given only the company name and the call date.

\paragraph{Robustness settings.}
The $\pm$3-year controls have no Item~4.02 observed within three years of their call, with observation ending in 2025; 12 of the 218 $\pm$3-year controls therefore have shorter follow-up. SPAC-related cases are warrant- or equity-classification restatements, filings from March to July 2021, or cases with the shell-company flag, which is also a baseline feature; the April 2021 SEC staff statement concerned how existing accounting guidance applies to SPAC warrants. In the out-of-time split, all test calls come after all training calls, and 3 of the 69 test sets have a call before the last training filing (29 October 2019). Without these 3 sets, $\Delta=+0.015$ [$-0.03$, 0.06]. The misstatement score of \citet{dechow2011predicting}, available for all three calls in 42 sets, reaches 0.560 [0.42, 0.69] on its own. Added to the baseline on all 109 sets with a missing-value indicator, it changes AUROC by $-0.005$.

\paragraph{Text and baseline scores.}
Adding the two models' standardized scores, instead of fitting one combined model, gives $+0.014$ [$-0.06$, 0.08]; this check was added without a plan. Within matched sets, the text and baseline scores have a rank correlation of 0.38 [0.25, 0.49]. A ridge regression that predicts the text score from the baseline's firm characteristics, within matched sets and under the same cross-validation, explains a small but non-random part of its variation ($R^2=0.05$; permutation $p=0.005$). The TF-IDF terms most associated with cases include trouble words such as investigation, legal, enforcement, and audit, and words from young industries such as cannabis and hydrogen; the terms most associated with controls include repurchases, billion, and record.

\paragraph{Conditional logit specifications.}
In the planned analysis of the 2015--2024 sets, with firm characteristics entered directly, the text coefficient has $p=0.045$. On all sets it has $p=0.075$. Across these two samples and the covariate specifications, $p$ ranges from 0.036 to 0.075.

\begin{figure}[h]
\centering
\includegraphics[width=\linewidth]{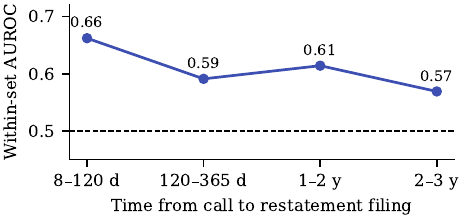}
\caption{Text-model performance by time between the call and the restatement filing. Within each window, scores are averaged over each firm's calls. A matched set is included when the case and at least one control have a call in that window (77, 93, 66, and 36 sets). Calls are scored by models that did not train on the firm.}
\label{fig:time}
\end{figure}

\paragraph{Time to disclosure.}
Figure~\ref{fig:time} shows the text score by time between the call and the filing. From the nearest to the farthest window, the estimates are 0.66, 0.59, 0.61, and 0.57. They vary without a consistent trend, and the windows contain different sets. On the 36 sets with calls in both periods, calls two to three years before the filing score 0.569 and the pre-filing calls 0.583. In 98 sets the case call itself presents results that were later restated, so the time to disclosure differs from the time since the accounting problem began.

\paragraph{Within-firm unusualness.}
A call's change is one minus the cosine similarity between its TF-IDF vector and the average of up to four earlier calls of the same firm within three years. The unusualness score divides this change, after subtracting the firm's average change, by the standard deviation of the firm's earlier changes, and requires at least five earlier calls. No firm had a zero standard deviation. The vocabulary and IDF weights are fitted without labels on all calls, including later ones, so the score describes calls retrospectively. At the pre-filing call the score reaches 0.439 [0.33, 0.57] on 49 sets. In an exploratory comparison added without a plan, on the 23 sets with enough history two to three years before the filing, the score separates cases from controls with AUROC 0.804 [0.65, 0.93] at those earlier calls, compared with 0.457 at the same sets' pre-filing calls (paired difference 0.35 [0.11, 0.57]). Whether this reflects the start of the accounting problem remains a hypothesis.

\paragraph{Explicit trouble disclosures.}
About 13\% of case calls and 2\% of controls mention investigations, audit-committee reviews, control weaknesses, or similar problems. These passages were identified by a language model (agreement with an independent model rater, Cohen's $\kappa=0.91$; \citealp{cohen1960kappa}) and deleted from the scored calls without refitting. Name masking refits the model on masked text.

\paragraph{Evasiveness.}
Qwen3.5-9B rated how directly each answer addressed the analyst's question in 5{,}850 question--answer exchanges. The resulting evasiveness score reaches 0.572 [0.49, 0.65] on 104 sets with Q\&A. The interval spans both chance and the 0.60 threshold, and agreement with a second model rater is $\kappa=0.37$.

\paragraph{Restatement reasons and era.}
Reasons were coded from the filings (agreement between a language-model pre-fill and independent model coders, $\kappa=0.82$). Among reasons with at least ten sets, warrant- or equity-classification restatements are the easiest to rank (0.82, 17 sets; 14 from the 2021 wave of warrant restatements). The text model's within-set AUROC is 0.70 for stock compensation (10 sets), 0.56 for revenue recognition (16), 0.50 for balance-sheet or cash-flow classification (14), and 0.54 for filings coded ``other'' (24). Excluding only warrant- or equity-classification restatements, the 2015--2024 sets still score 0.648. The broader SPAC-related exclusion in Table~\ref{tab:robust} also removes 2021-wave filings and flagged firms from both eras. The era difference is confounded with corpus: within one corpus it is 0.19 [$-0.04$, 0.40].

\end{document}